\documentclass{article}
\usepackage{PRIMEarxiv}

\usepackage[colorlinks,urlcolor=blue,linkcolor=blue,citecolor=blue]{hyperref}
\usepackage{color,array}
\usepackage{graphicx}

\usepackage{amsmath} 
\usepackage{amssymb}  
\usepackage{color}
\usepackage{url}  
\usepackage{textcomp}
\usepackage{graphicx}          
\usepackage[dvips]{epsfig}

\usepackage{caption}
\usepackage{subcaption}

\usepackage{multicol}

\newcommand{\NN}{\mathcal{N}} 
\newcommand{\D}{\mathcal{D}} 
\newcommand{\nsamp}{n}
\newcommand{\batchsize}{b}
\newcommand{\seqlen}{m}

\newcommand{\nin}{{n_u}} 
\newcommand{\ny}{{n_y}} 
\newcommand{\nx}{{n_x}}

\newcommand{\norm}[1]{\left\lVert#1\right\rVert}
\newcommand{\simul}[1]{{#1}^{\rm sim}}
\newtheorem{remark}{Remark}

\usepackage{authblk}

\title{Learning neural state-space models: do we need a state estimator?}

\author{Marco Forgione}
\author{Manas Mejari}
\author{Dario Piga}
\affil{IDSIA Dalle Molle Institute for Artificial Intelligence USI-SUPSI, Via la Santa 1, CH-6962 Lugano-Viganello, Switzerland.}

\begin{document}

\maketitle

\begin{abstract}
In recent years, several algorithms for system identification with neural state-space models have been introduced. Most of the proposed approaches are aimed at reducing the computational complexity of the learning problem, by splitting the optimization over short sub-sequences extracted from a longer training dataset. Different sequences are then processed simultaneously within a minibatch, taking advantage of modern parallel hardware for deep learning. An issue arising in these methods is the need to assign an initial state for each of the sub-sequences, which is required to run simulations and thus to evaluate the fitting loss. In this paper, we provide insights for calibration of neural state-space training algorithms based on extensive experimentation and analyses performed on two recognized system identification benchmarks. Particular focus is given to the choice and the role of the initial state estimation. We demonstrate that advanced initial state estimation techniques are really required to achieve high performance on certain classes of dynamical systems, while for asymptotically stable ones basic procedures such as zero or random initialization already yield competitive performance.
\end{abstract}


{\section{Introduction}\label{sec:introduction}}

While neural networks have been employed for dynamical systems modeling since (at least) the early 90s of the last century \cite{chen1990non}, the recent impressive achievements of deep learning in diverse domains \cite{schmidhuber2015deep} has sparked renewed interest
from the system identification community.

Nowadays, the combined availability of flexible open-source software for deep learning \cite{pytorch} and cheap hardware supporting massive parallel computing enables rapid prototyping and testing of different neural architectures. Leveraging these tools, tailor-made neural model structures designed for dynamical system modeling have been developed. To cite a few examples, state-space models embedding \emph{a priori} system knowledge, and using neural networks to describe the uncertain system components have been showcased in \cite{forgione2020model}. Novel deep-learning architectures embedding linear transfer function as elementary building blocks have been introduced in \cite{forgione2021dynonet}, along with a specialized and efficient implementation of the back-propagation computations for the transfer-function operator. Finally, 
\cite{iacob2021deep, lusch2018deep} propose modeling approaches based on Koopman operator theory. The system state is lifted to a high-dimensional space where the dynamics evolve linearly. This linear dynamics, along with the non-linear state transformation (parametrized as a neural network) 
are jointly learned with deep-learning tools.

System identification researchers have also worked towards the development of efficient algorithms to fit these neural model structures to experimental data. In particular, when the training dataset consists of a single, long (w.r.t. the system's time constants) sequence, standard simulation-error minimization has been show to be either computationally inefficient or even infeasible \cite{forgione2020model, ribeiro2020smoothness}. Conversely, one-step-ahead approaches may be sensitive to measurement noise, and in general they are less effective in producing reliable long-term predictions.

An intermediate approach has thus emerged, where an $\seqlen$-step-ahead prediction error is minimized over \emph{sub-sequences} extracted from the training dataset
\cite{forgione2020model, beintema2021nonlinear, masti2021learning}. 
The length $m$ of the considered sub-sequences is an important hyper-parameter of the training algorithm.
This approach presents several advantages. First, from a computational perspective, several sub-sequences can be processed simultaneously, thus taking full advantage of modern parallel hardware \cite{forgione2020model}. Second, as demonstrated in \cite{ribeiro2020smoothness}, the optimization problem to be solved when minimizing the prediction error over a  very long horizon may be ill-posed, notably in the case of unstable and chaotic dynamics. The multi-step-ahead prediction method enables controlling this effect through tuning of the sub-sequence length $\seqlen$.
Lastly, the ultimate aim of modeling is often to generate accurate predictions over a specific time horizon. Notably, this occurs when the aim is to design a model predictive controller \cite{borrelli2017predictive}. In this case, $m$-step ahead prediction error minimization with $\seqlen$ equal to the envisaged prediction horizon of the controller matches the final goal closely.

A difficulty in the implementation of $\seqlen$-step-ahead prediction-error minimization is the need to start simulations at different time instants along the full training trajectory.
For state-space models, in particular, we need to assign an \emph{initial state} to all the sub-sequences considered during training. Thus,
in \cite{forgione2020model}, the full state trajectory is considered as an optimization variable to be estimated along with the parameters of the model dynamics.
In \cite{beintema2021nonlinear}, an \emph{encoder} network is used as a deadbeat state estimator, and its parameters are trained together with the ones of the model to minimize the $m$-step-ahead prediction error. The work may be seen as an extension (to the $m$-step-ahead case) of the one-step method previously described in \cite{masti2018learning}.

In the above-mentioned papers, neural state-space model learning approaches are introduced and tested on a limited number of configurations. The contributions do not include extensive experimentations exploring the algorithm hyper-parameters in a systematic fashion, nor a statistical analysis of their main effects and interactions.

The aim of this paper is to shed light on the role of different user choices and algorithm settings for the training of state-space neural models through $m$-step-ahead simulation error minimization, with a particular focus on the initial state estimation strategy.
Compared to the previous contributions, we also report the performance obtained with \emph{dummy} estimation strategies such as \emph{random} and \emph{zero} state initialization, to be considered as baselines. Interestingly, these two baselines are shown to be competitive on the Wiener-Hammerstein benchmark \cite{ljung2009wiener} considered for neural state-space identification in \cite{beintema2021nonlinear}. We then perform another extensive analysis involving the pick-and-place machine benchmark \cite{juloski2004pickplace}, which (to the best of our knowledge) has not been previously considered to asses neural state-space models. On the pick-and-place benchmark, we show that more advanced strategies for initial state estimation are really required to attain high fitting performance. With these strategies, we have obtained the highest model fit on this benchmark w.r.t.  the ones reported in the literature thus far. 

Finally, we analyse and infer from the numerous results obtained on the considered benchmarks taking into account properties of the underlying systems, and provide guidelines for system identification practitioners.

\section{Settings}
\subsection{Notation}
Consider an $\nsamp$-length sequence of variables $z=\{z_0, z_1, \dots, {z_{\nsamp-1}}\}$, where each element $z_i$ is either scalar- or vector-valued. We adopt (python-like) \emph{slicing} notation to address a portion of the sequence, \emph{i.e.}, for integers $h,k$, with $h \leq k \leq \nsamp -1$, 
$z_{h:k}=\{z_h, z_{h+1}, \dots, z_{k-1}\}$ denotes the sub-sequence with start index $h$ (included)  and stop index $k$ (excluded).


\subsection{Dataset and model structure}
We are given a dataset $\D=(u_{0:\nsamp}, y_{0:\nsamp})$ containing $\nsamp$ input samples $u_i \in R^{\nin}$ and (possibly noisy) output samples $y_i \in 
R^{\ny}$, collected from a dynamical data-generating system $\mathcal{S}$. Our goal is to estimate a  model $M$ of the unknown dynamics of $\mathcal{S}$ using the dataset $\D$.

We consider the following model structure:
 \begin{subequations}
  \label{eq:ss_model}
 \begin{align}
  x_{k+1} &= \NN_f(x_k, u_k; \theta) \label{eq:ss_model_a} \\
  y_{k} &=  \NN_g(x_k; \theta), \label{eq:ss_model_b} 
  \end{align}
\end{subequations}
where $\NN_f$ and $\NN_g$ are feed-forward neural networks having compatible dimensions, $x_k \in \mathbb{R}^{\nx}$ is the state at time $k$ and $\theta
\in \mathbb{R}^{n_\theta}$ is a vector of parameters to be identified from data. Overall, \eqref{eq:ss_model} defines a neural state-space model structure, 
often referred to as \emph{recurrent neural network}.
\subsection{Full simulation error minimization}
To estimate the model parameters $\theta$, a straightforward approach is to minimize the simulation error norm with respect to both $\theta$ and the initial state $x_0$:
\begin{equation}
\label{eq:loss_simerr}
 \hat \theta, \hat x_0 = \arg \min_{\theta, x_0} 
 \sum_{k=0}^{\nsamp-1} \norm{y_k - \simul{y}_k}^2,
\end{equation}
where $\simul{y}_k$ is the output obtained by iterating (thus, simulating) the model dynamics \eqref{eq:ss_model} over $\nsamp$ time steps.
From a theoretical perspective, minimization
of \eqref{eq:loss_simerr} corresponds, in the case of additive white Gaussian noise on the measured outputs $y_k$, to the \emph{maximum likelihood} estimation
of $\theta$ and $x_0$. Thus, in principle, the methodology enjoys desirable properties, \emph{e.g.}, consistency and asymptotic optimality \cite{hastie2009elements}. However, from a practical perspective, the method suffers from two severe limitations. 
First, repeated (forward and backward) evaluation of the full simulation error minimization is often computationally heavy for long training sequences, and it offers limited opportunities for parallelization due to the \emph{sequential} nature of the simulation over time \cite{forgione2020model}. Second, the simulation error norm may be numerically hard to optimize (\emph{i.e.}, the conditioning of the optimization problem is often poor), notably in the case of unstable or chaotic underlying dynamics \cite{ribeiro2020smoothness}.


\subsection{Truncated simulation error minimization}
To overcome the above limitations, it is possible to minimize the simulation error over shorter sequences extracted from the dataset.
Considering all possible $\seqlen$-length contiguous sequences from $\D$, the optimization problem becomes:
\begin{equation}
\label{eq:loss_with_states}
 \hat \theta,  \hat x_{0:\nsamp-\seqlen} = \arg \min_{\theta, x_{0:\nsamp-\seqlen}} 
 \sum_{i=0}^{\nsamp - \seqlen - 1} 
 \overbrace{\sum_{j=0}^{\seqlen-1} \norm{y_{i+j} - \simul{y}_{i+j|i}}^2}^{=J_i},
\end{equation}
where $\simul{y}_{i+j|i}$ is the simulated output at time step $i+j$ obtained by initializing the simulation 
at time $i$ with state $x_{i}$. Computation of the loss \eqref{eq:loss_with_states} can be easily parallelized for the $(\nsamp - \seqlen)\seqlen$ different $\seqlen$-length sequences as the terms $J_i$ do not depend on each other and can be processed independently (possibly exploiting parallel hardware).
However, the formulation requires the inclusion of an optimization 
variable $x_{0:\nsamp-\seqlen}$ representing the state sequence over $\nsamp-\seqlen$ time steps.
The larger number of optimization variables may then lead to an increased variance of the identified parameters $\hat \theta$, and thus make the algorithm less sample-efficient. 

To counter the variance increase, an effective regularization approach enforcing
\emph{compatibility} between the estimated state sequence $\hat x_{0:\nsamp-\seqlen}$ and the dynamics \eqref{eq:ss_model} was introduced in \cite{forgione2020model}.
However, the approach in~\cite{forgione2020model} requires careful tuning of the trade-off loss (fitting vs. regularization), and it is not further discussed here.

\subsection{State estimator}
Instead of treating the full state sequence as an optimization variable, it is possible to learn a \emph{state estimator} that maps previous
input/output data samples to the current (estimated) state. We consider, in particular,  \emph{deadbeat} estimators which generate an estimate $\hat x_i$ of the state
at time $i$ from a \emph{finite} window of past input/output measurements of length $m_e$:
\begin{equation}\label{eqn:estimator}
 x_{i} = \NN_e(u_{i-\seqlen_e:i}, y_{i-\seqlen_e:i}; \phi).
\end{equation}
Exploiting such a deadbeat structure, it is still possible to define a multi-step cost function that can be split into independent contributions for the sequences:
\begin{subequations}
\label{eq:loss_with_estimator}
\begin{align}
 \hat \theta, \hat \phi &= \arg \min_{\theta, \phi} 
 \sum_{i=0}^{\nsamp - \seqlen - 1} \sum_{j=\seqlen_e}^{\seqlen -1} \norm{y_{i+j} - \simul{y}_{i+j|i}}^2\\
  \hat x_{i+\seqlen_e} &= \NN_e(u_{i:i+\seqlen_e}, y_{i:i+\seqlen_e}; \phi),
 \end{align}
\end{subequations}
with $\seqlen = \seqlen_e + \seqlen_f$. 
Note that the $m$-length training sequence starting at time index $i$ may be seen as divided into an initial $\seqlen_e$-lenght portion used just to reconstruct the state at time $i+\seqlen_e$, and a trailing $m_f$-length portion used to compute the fitting loss.

The strategy above may be more attractive than an explicit estimation of the whole state sequence $x_{0:\nsamp-\seqlen}$ (as done in \eqref{eq:loss_with_states}) when the training dataset is relatively large and the dimension of the parameter vector $\phi$ required to describe the estimator is (expected to be) smaller than the sequence $x_{0:\nsamp-\seqlen}$ itself.

\begin{remark} \label{rem:min_seq_est_len_1}
In \cite{verdult2004least}[\textbf{Lemma 1}], exact state reconstruction is shown to be possible (in the noise-free case) by choosing $\seqlen_e \geq \nx$.  Specifically, the Lemma states that: if the data-generating system is of the form~\eqref{eq:ss_model}, and it  is strongly locally observable, then under mild conditions, for all $\seqlen_e \geq \nx$ there exists a smooth function $\gamma: \mathcal{U} \times \mathcal{Y} \rightarrow \mathcal{X}$, where $\mathcal{U} \subset \mathbb{R}^{\nin \seqlen_e},  \mathcal{Y} \subset \mathbb{R}^{\ny \seqlen_e},  \mathcal{X} \subset \mathbb{R}^{\nx}$,  such that the state at time step $i$ is given by $ x_{i} = \gamma(u_{i-\seqlen_e:i}, y_{i-\seqlen_e:i})$. 
In this sense, the estimator  $x_{i} = \NN_e(u_{i-\seqlen_e:i}, y_{i-\seqlen_e:i}; \phi)$ given in \eqref{eqn:estimator} can be seen as a machine-learning surrogate of the smooth function
$\gamma$. Classic universal approximation theorems for neural networks~\cite{cybenko1989approximation} 
guarantee that, for a sufficiently complex architecture, the 
approximation error of $\NN_e$ can be made arbitrarily small.
\end{remark}

\begin{remark}
In this paper, we restrict ourselves to \emph{causal} estimators producing the state estimate 
$\hat x_i$ at time $i$ based on \emph{previous} input/output samples $u_{i-\seqlen_e:i}, y_{i-\seqlen_e:i}$. In principle, current and future input/output samples could also be considered to construct the state estimate. We leave analysis and experimentation with non-causal state reconstruction techniques as future work.
\end{remark}

\subsubsection{State estimator structure}
\label{subsubsec:state_estimation_structure}
In principle, the state estimator $\NN_e$ could be designed using traditional techniques from the control literature \cite{simon2006optimal} (\emph{e.g.}, Luenberger, Kalman, moving horizon \emph{etc.}). However, as argued in \cite{beintema2021nonlinear}, standard feed-forward (FF) neural networks (which are  easily implemented and integrated in deep-learning software frameworks) may be applied for this task as well.

Building upon \cite{beintema2021nonlinear}, in this paper we also explore the case where $\NN_e$ is parametrized as a long-short-term memory (LSTM) network  processing the 
$m_e$ input/output samples sequentially, and we provide comparative results with the FF case. 

Furthermore, we include in our experiments  two ``dummy'' estimators called ZERO and RAND, to be used as baseline for the most advanced ones (FF, LSTM).
The ZERO estimator uses the $\seqlen_e$ input/output estimation samples to start a simulation with the initial state set equal to 0. The simulated state at the last step of the estimation sequence is then used as initial state for fitting. Similarly, the RAND estimator initializes the simulation in the estimation sequence with a random value extracted from a standard Gaussian distribution and generates the initial state for the optimization in fitting sequence as the state obtained at the last step of the estimation horizon.

\subsection{Implementation aspects}
We briefly state here implementation aspects followed in this paper. The presentation is
rather concise as they are in line with standard ``good practice'' for deep learning, see \cite{michelucci2018applied}.

Optimization of the loss \eqref{eq:loss_with_estimator} is performed by stochastic gradient descent over (mini)batches of sequences extracted from the training dataset. At each iteration of the optimization loop, the following minibatch loss is computed: 
\begin{subequations}
\begin{align}
\label{eq:loss_with_estimator_minibatch}
 \tilde J(\theta, \phi) &= \frac{1}{\batchsize\seqlen}\sum_{s=0}^{\batchsize -  1}
 \sum_{j=\seqlen_e}^{\seqlen-1} \norm{y_{i_s+j} - \simul{y}_{i_s+j|i_s}}^2\\
 \hat x_{i_s+\seqlen_e} &= \NN_e(u_{i_s:i_s+\seqlen_e}, y_{i_s:i_s+\seqlen_e}; \phi),
 \end{align}
\end{subequations}
where the batch size $\batchsize$ is the number of sequences included in the batch and  $i_s \in \{0,\ldots, \nsamp\! - \seqlen\! - 1\}$  is a random integer defining the starting index of the sequence $s$, for $s=0,1, \dots, \batchsize-1$.

The gradients of \eqref{eq:loss_with_estimator_minibatch} with respect to the optimization variables $\theta$ and $\phi$ are computed with reverse-mode automatic differentiation, as implemented in standard deep learning software \cite{pytorch} and used for iterative gradient-based optimization. 
For numerical optimization, the Adam algorithm \cite{kingma2014adam} is used, since it is generally recognized as more iteration-efficient than plain gradient descent, while not 
introducing significant additional computational burden.

A 20\% portion of the training data is used as hold-out validation dataset. For each configuration, the model having the lowest validation loss over the optimization iterations is saved. 

Instead of a fixed number of iterations, we run the training
 algorithm for a fixed amount of \emph{time}. This allows for a fair comparison among algorithm configurations characterized by widely varying computational cost per iteration.

\section{Experiments}
The methodologies described in the paper are evaluated on two standard benchmarks for system identification, namely the Wiener-Hammerstein circuit \cite{ljung2009wiener} and the pick-and-place machine \cite{juloski2004pickplace}. For the two examples, extensive experimental campaigns exploring the effect of different training algorithm settings
on the obtained model's performance are carried out. %

The trained models are then evaluated in terms of their FIT index on a test dataset:
\begin{equation}
\label{eq:fit_index}
\mathrm{FIT} = 100 \times \left(1- \frac{\sqrt{\sum_{t=0}^{\nsamp-1} \left({y}_t -  {y}^{\rm sim}_t\right)^2} }  
{\sqrt{\sum_{t=0}^{\nsamp-1} \left({y}_t -  {\overline{{y}}}\right)^2}}\right) (\%),
\end{equation}
where ${y}$ is the measured output; ${y}^{\rm sim}$ is the model's simulated output;  
and $\overline{{y}}$ is the mean value of $y$, i.e. $\overline{{y}} = \frac{1}{\nsamp} \sum_{t=0}^{\nsamp-1} {y}_t$.
We then inspect the results of our experimental campaigns using standard tools for statistical analysis \cite{montgomery2017design}, taking the settings of the training algorithm as controlled \emph{factors} and the model's $\mathrm{FIT}$ index in the test dataset as measured \emph{response}.

All the codes required to reproduce our results are available in the GitHub repository \url{https://github.com/forgi86/sysid-neural-estimator}. The user may also 
extend the analysis to additional factors (and new benchmarks) with minor modifications to the provided software.

Computations are performed on a deep-learning server of the IDSIA laboratory equipped with 2 64-core AMD EPYC 7742 Processors, 256 GB of RAM, and 4 Nvidia RTX 3090 GPUs. In all the experiments, the hardware resources of the server have been limited to 10 CPU threads and 1 GPU.

\subsection{Wiener-Hammerstein circuit}
This benchmark was introduced in \cite{ljung2009wiener} and it contains real measurement from an electronic circuit which implements a Wiener-Hammerstein (WH) dynamics. 
Input and output samples are collected at a constant frequency of 51200~Hz. The training and test datasets contain 100000  and 87000 samples, respectively.

In line with \cite{beintema2021nonlinear}, the number of states of the model  \eqref{eq:ss_model} is set to $n_x=6$ and the neural networks $\NN_f$ and $\NN_g$ defining the state update and output mapping are chosen having a single hidden layer with 15 nodes and $\mathrm{tanh}$ activation function, plus a direct linear term from input to output. 
We consider in our experiments the cases of state estimators $
\NN_e$ with FF, LSTM, ZERO, and RAND structure (see Section~\ref{subsubsec:state_estimation_structure}). For the FF estimator,
we use a single-hidden-layer neural architecture with 15 nodes and a direct input/output linear term. For the LSTM estimator, we employ 
a standard implementation (as available in PyTorch) with a single hidden layer and 16 nodes. The LSTM processes the samples in the estimation sequence forward in time, and its output at the last time step defines the initial state for fitting. Finally, the learning rate of Adam is fixed to $1\cdot 10^{-3}$. 

The factors explored in the experimental campaign are  reported in Table \ref{tab:parameters_wh}, together with their considered levels. The estimator type (est\_type), estimation sequence length $\seqlen_e$ (seq\_est\_len) and fitting sequence length $\seqlen_f$ (seq\_fit\_len) are included, as well as the batch size $\batchsize$ (batch\_size) and the total training time (train\_time).
A \emph{full factorial} experimentation including
all combinations of the algorithm settings has been performed, totaling 768 training runs which required approximately 17 days.\footnote{Note that the total experimental time is determined \emph{a priori} as we execute the training algorithm for a fixed amount of time, and not for a fixed number of iterations.}
\begin{table}
    \centering
    \begin{tabular}{|l||l|}
    \hline
    Factor &  Values \\
    \hline
Estimator type (\texttt{est\_type}) & FF, LSTM, ZERO \\
& RAND \\
\hline 
Training time (s) (\texttt{max\_time}) & $300$, $1800$, $3600$ \\
\hline 
Size $b$ of training (mini)batch    (\texttt{batch\_size}) & $32$, $128$, $512$, $1032$ \\ 
\hline
Fitting seq. length (\texttt{seq\_fit\_len}) &  $40$, $80$, $160$, $320$ \\
\hline
Estimation seq. length $m_e$  (\texttt{seq\_est\_len}) &  $10$, $20$, $40$, $80$ \\
\hline 
    \end{tabular}
    \vspace{0.2cm}
    \caption{WH circuit: Parameter configuration of the experiments.}
    \label{tab:parameters_wh}
\end{table}

The top-5 configurations in terms of FIT index over the test dataset are reported in Table~\ref{tab:top_config_wh}.
\begin{table}[t!]
    \centering
    \begin{tabular}{|c|c|c|c|c|}
    \hline
    \texttt{est\_type}  & \texttt{batch\_size} & \texttt{seq\_fit\_len} & \texttt{seq\_est\_len} & \texttt{FIT} \\
    \hline
LSTM  & 128 & 80 & 40 & 98.96 \% \\
RAND  & 512 & 80 & 40 & 98.89 \% \\
ZERO & 512 & 40  & 80 & 98.87 \% \\
RAND  & 128 & 40 & 40 & 98.82 \% \\
RAND & 1024 & 80 & 80 &  98.81 \% \\
\hline
\end{tabular}
\vspace{0.2cm}
    \caption{WH circuit: Top-$5$ configurations. For all the configurations, train\_time=3600~s.}
    \label{tab:top_config_wh}
\end{table}
All these configurations are characterized by the longest training time (train\_time=3600~s). Interestingly, the best result is obtained with an LSTM estimator, and the 4 other instances in the top-5 list are based on dummy (ZERO/RAND) estimators. 

\begin{figure}
    \centering
    \includegraphics[width=.6\linewidth]{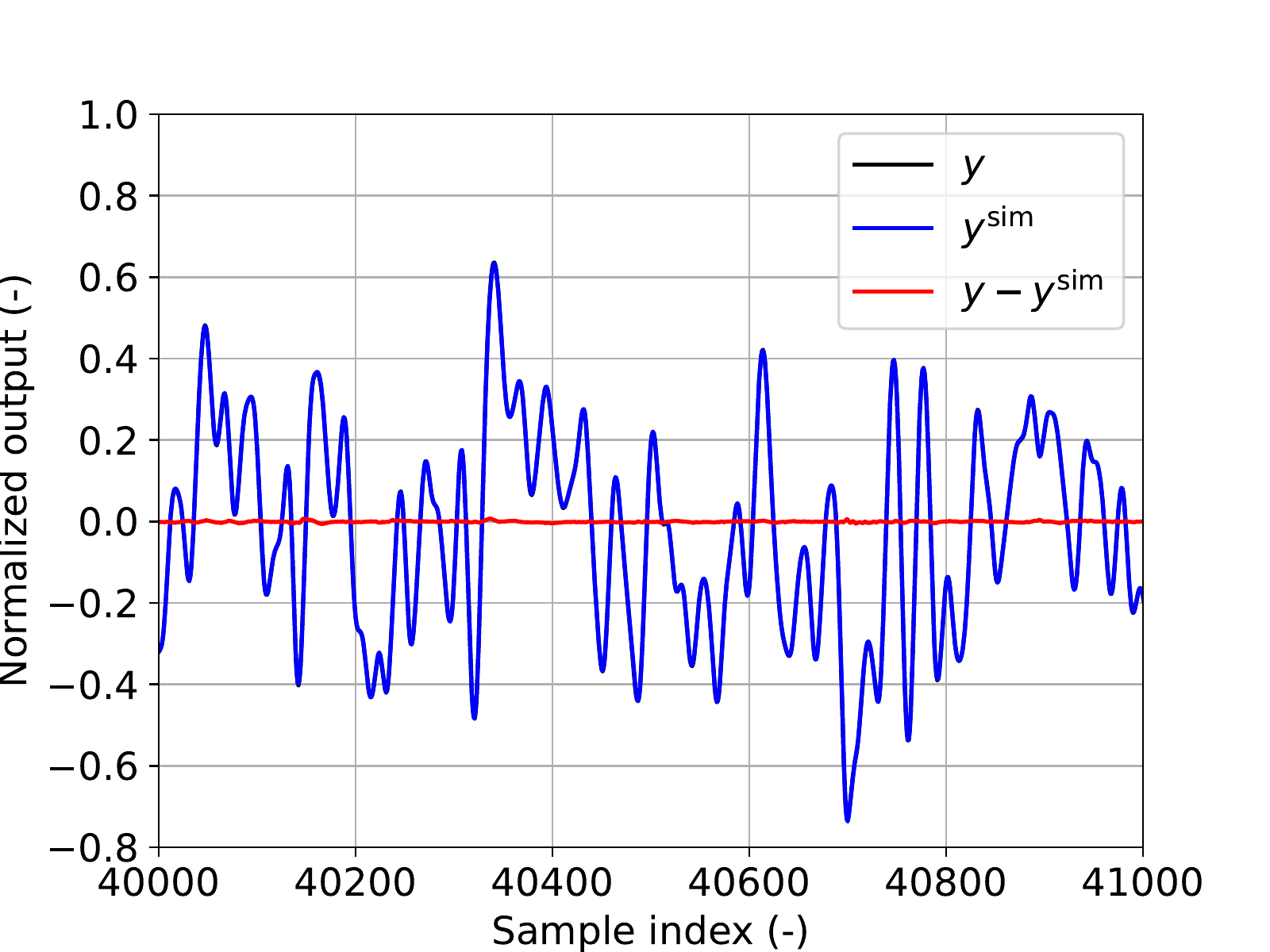}
    \caption{WH circuit: Measured output $y$, simulated output $y^{\rm sim}$, and error signal $y - y^{\rm sim}$ on a portion of the test dataset.}
    \label{fig:wh_best_timetrace.pdf}
\end{figure}

The measured output $y$, the output $y^{\rm sim}$ simulated from the best obtained model, and the error signal $y - y^{\rm sim}$ are shown in Fig.~\ref{fig:wh_best_timetrace.pdf}. For better visualization, only a short portion of the test dataset is displayed.

\begin{figure}
    \centering
    \includegraphics[width=.6\linewidth]{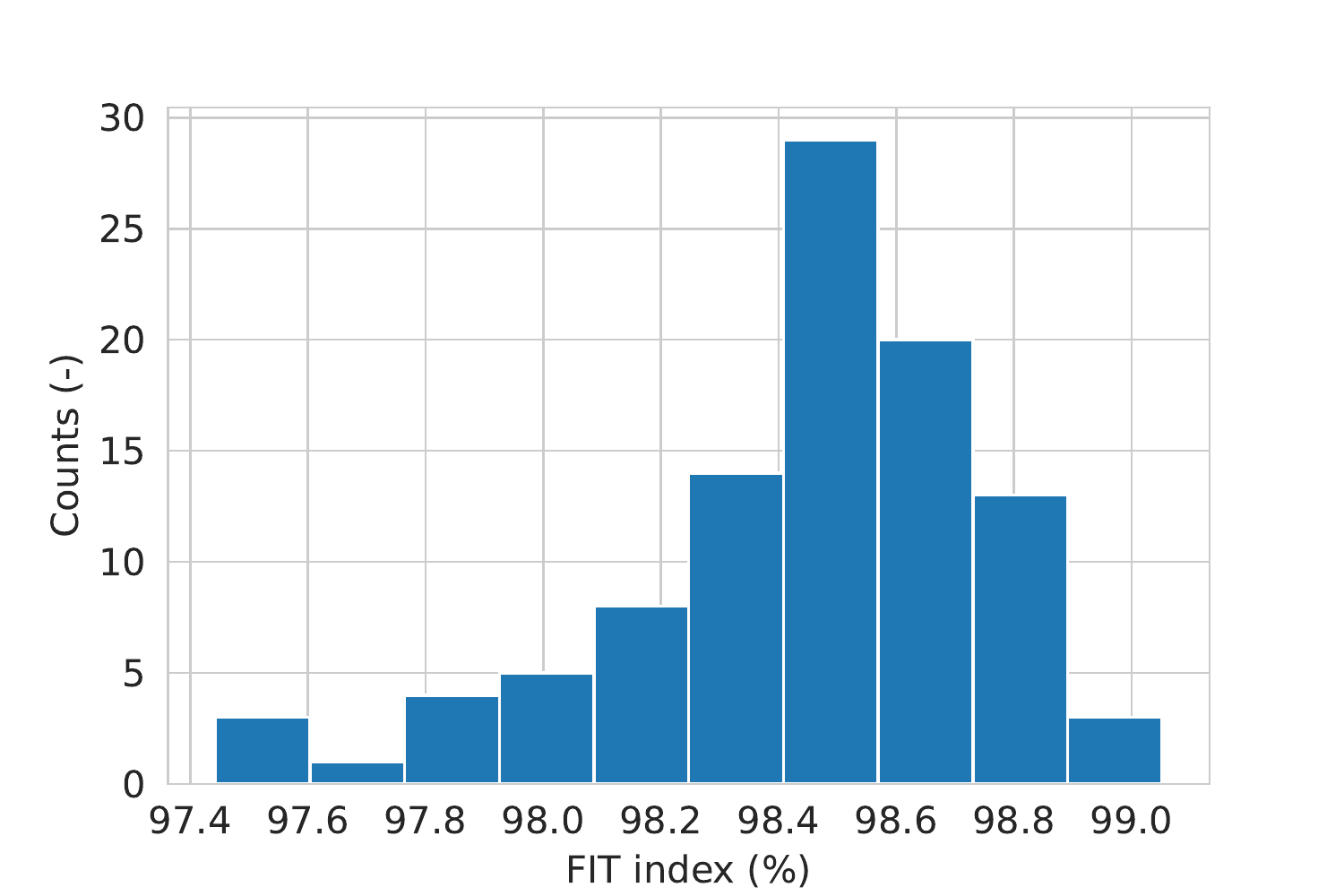}
    \caption{Histogram of the test FIT index achieved by repeating the best configuration 100 times.}
    \label{fig:wh_best_hist}
\end{figure}

In order to assess the intrinsic variability of the training algorithm, 100 repetitions of the best configuration with different random initializations of the model (and estimator) parameters have been executed.\footnote{Parameter initialization is the main source of non-determinism in the training algorithm.} The histogram of the FIT index achieved over the test dataset by the 100 trained models is shown in Fig. \ref{fig:wh_best_hist}.
The empirical mean and standard deviation of the 100 FIT values are 98.44\% and 0.32\%, respectively.
In this sense, FIT index differences larger than $3 \times 0.32\% \approx 1.0\%$ between different configurations are expected to be statistically significant.

\begin{figure}
    \centering
    \includegraphics[width=.7\columnwidth]{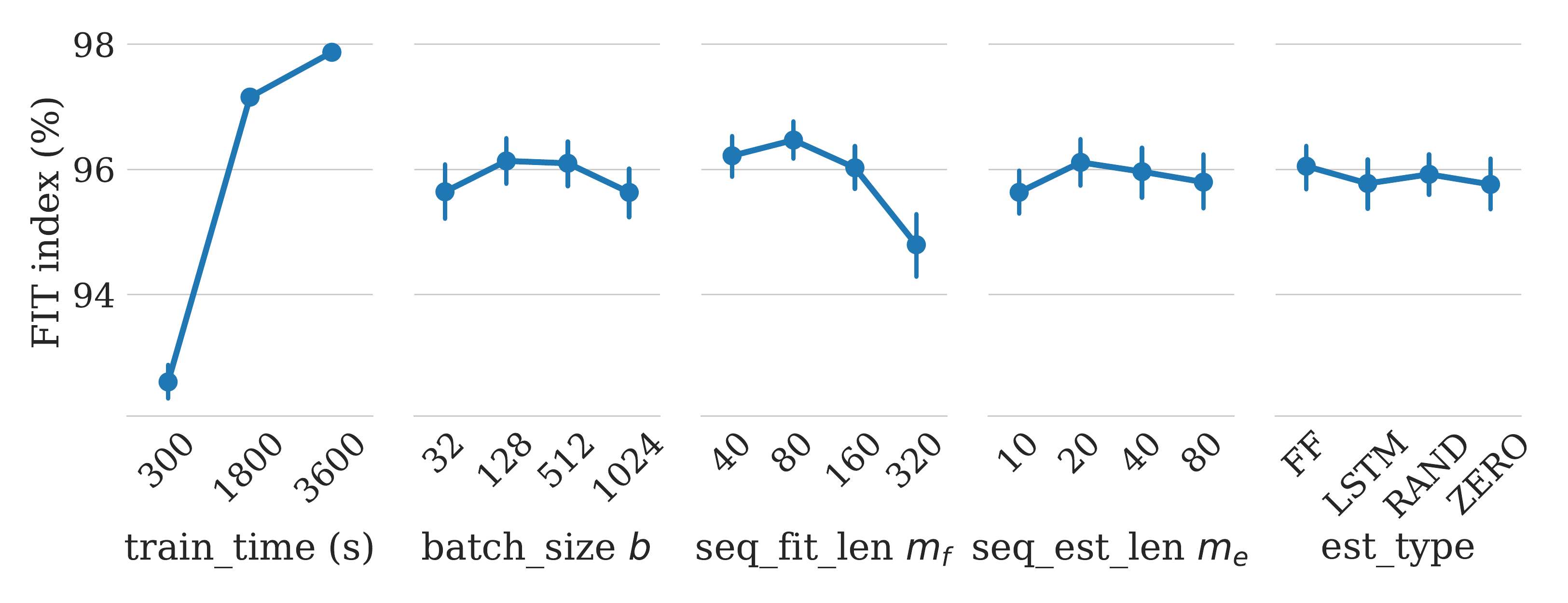}
    \caption{WH circuit: Main effects (and 95\% confidence intervals) of different factors.}
    \label{fig:main_effects_wh}
\end{figure}

In Fig.~\ref{fig:main_effects_wh}, the main effects of the factors are illustrated. For each level of a factor, the graph shows the average (w.r.t. all the other factor values) of the FIT index as a round dot, and a 95\% confidence interval as a vertical band.

The training time has by far the largest impact and it has a positive effect on the outcome, in line with our expectation. The fitting sequence length $\seqlen_f$ has also a significant effect, with a negative impact for the largest considered value $\seqlen_f=320$.
The influence of the state estimator choice (est\_type) is not visible at this level.

\begin{figure}
    \centering
    \includegraphics[width=.7\columnwidth]{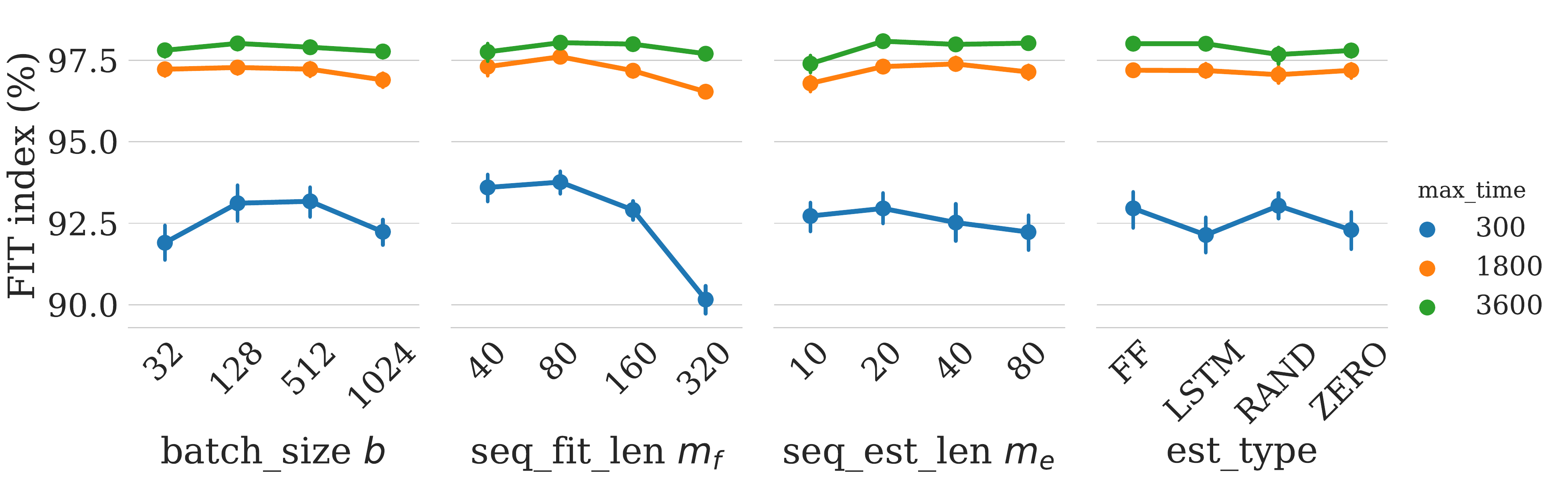}
    \caption{WH circuit: Interactions between train\_time and other factors.}
    \label{fig:train_time_interactions_wh}
\end{figure}

In Fig.~\ref{fig:train_time_interactions_wh}, the interactions between the training time (which is the dominant factor of the main effect analysis) and all the other factors are highlighted.
From the figure, it is evident that the negative effect of the longest fitting sequence length (seq\_fit\_len $\seqlen_f=320$) is particularly strong in conjunction with  the shortest training time (train\_time=$300~s$), and it is almost negligible in the case of the largest training time (train\_time=$3600~s$). Thus, the lower performance is likely related to the increasing computational cost per optimization iteration, which in turn resulted in a lower number total iterations executed in the fixed time budget.

\begin{figure}
    \centering
    \includegraphics[width=.7\columnwidth]{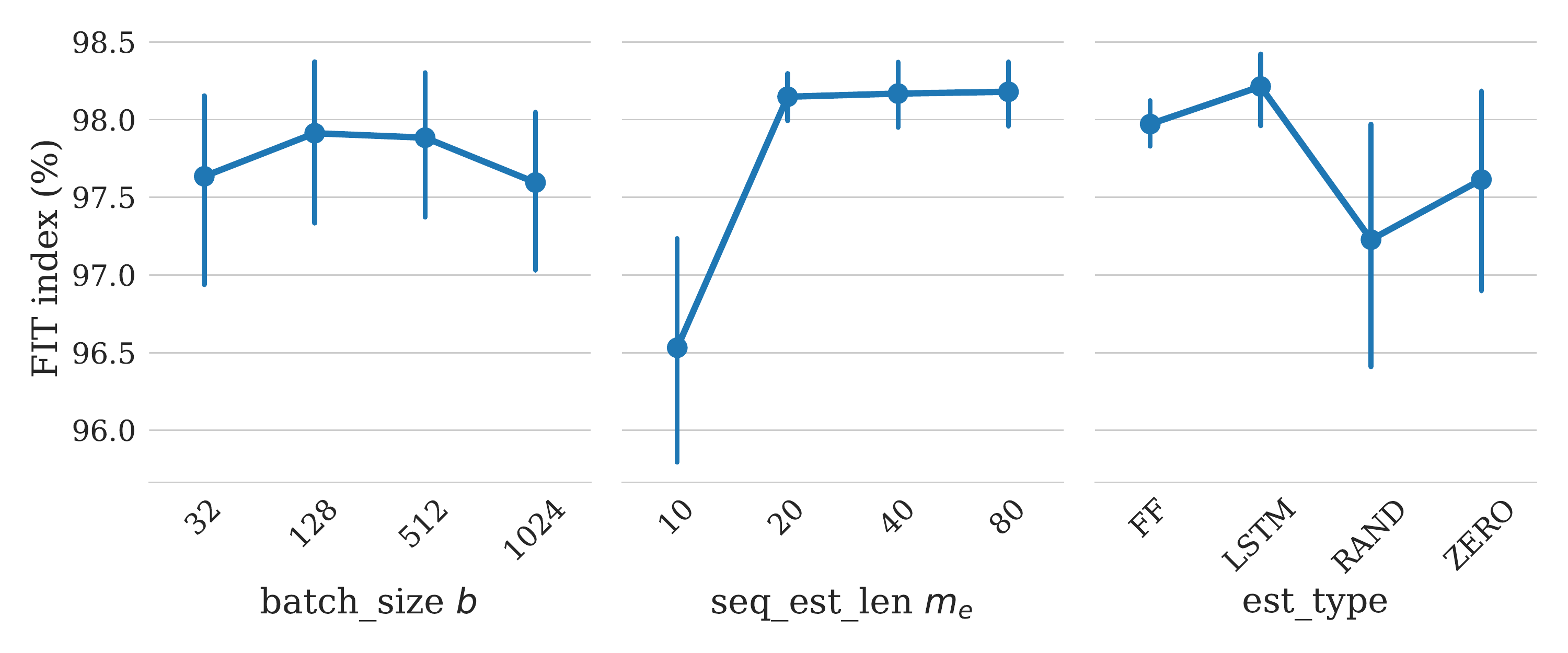}
    \caption{WH circuit: Main effects of factors restricted to train\_time=3600~s, seq\_fit\_len $m_f=40$.}
    \label{fig:main_effects_restricted_wh}
\end{figure}

In this benchmark, the effect of the state estimator is only visible by restricting the analysis to the shortest value of seq\_fit\_len $m_f=40$ and a longest training time (train\_time=3600~s).
The main effects of the remaining factors are visualized in Fig. \ref{fig:main_effects_restricted_wh}, where it appears that higher average performance with lower variance is associated with the experiments
where LSTM or FF state estimators are used. Remarkably, a significantly low average FIT  and high variability is also associated with the shortest estimation sequence length $\seqlen_e=10$.

\begin{figure}
    \centering
    \includegraphics[width=.5\columnwidth]{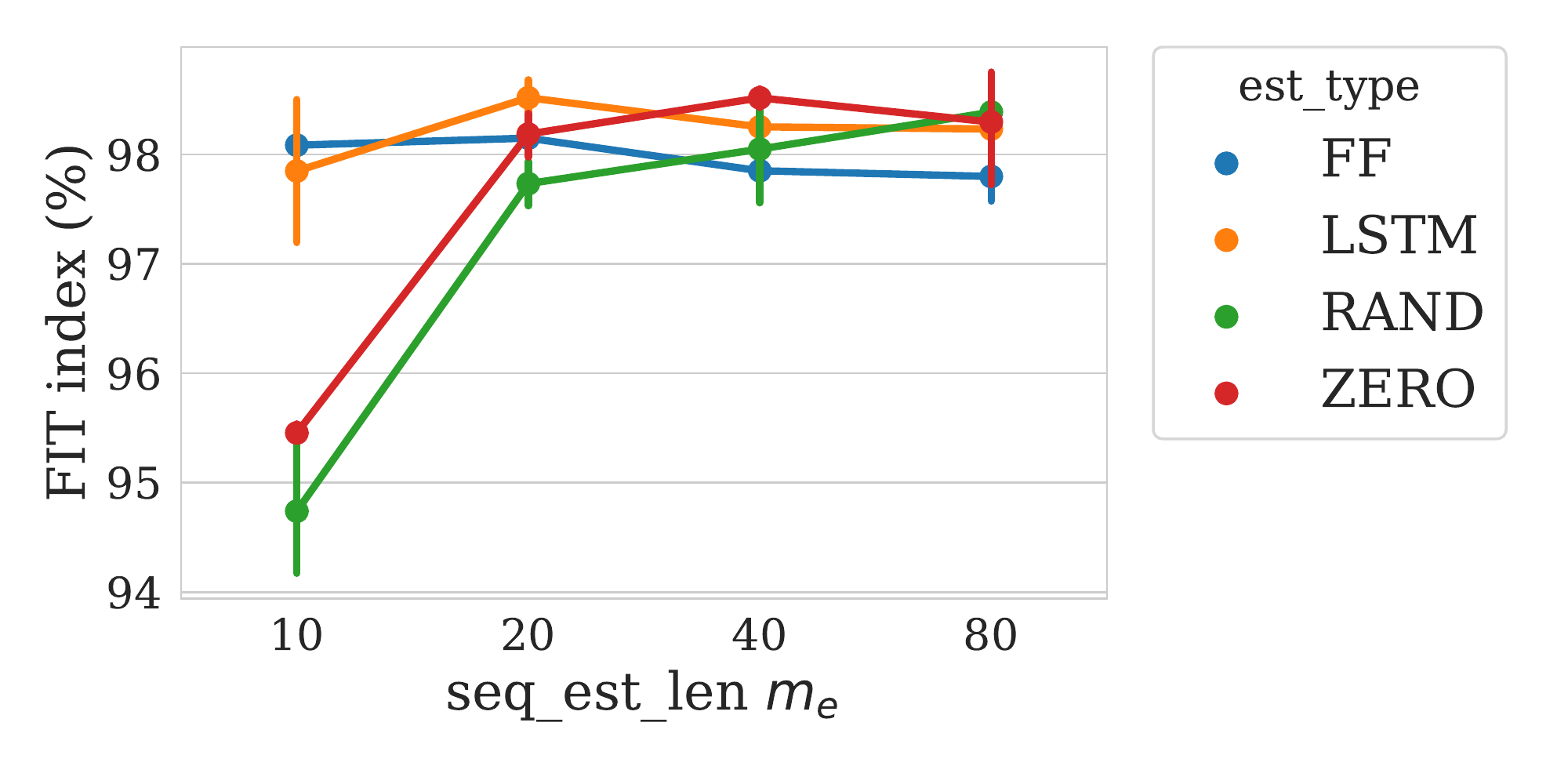}
    \caption{WH circuit: Interaction between factors seq\_est\_len and est\_type for train\_time=3600~s and seq\_fit\_len $m_f=40$.}
    \label{fig:interactions_restricted_wh}
\end{figure}

Further insight is obtained by looking at the interaction between est\_type and seq\_fit\_len in the 
same restricted set of configurations (Figure \ref{fig:interactions_restricted_wh}). 
From this figure, it is clear that a significantly lower performance is associated with experiments where the estimation sequence is short 
(seq\_est\_len $\seqlen_e=10$) and a dummy estimator (ZERO/RAND) is used. 

These observed results are compatible with the following reasoning: the Wiener-Hammerstein benchmark has an asymptotically stable dynamics, and state estimation may be performed simply by simulating an accurate model for a sufficiently long time horizon, starting from \emph{any} initial condition (including zero and random values).
Only for very short values of $\seqlen_e$ (w.r.t. the system dynamic's time constant), the dummy
estimation methods ZERO and RAND are not sufficiently accurate, resulting in a lower overall fitting performance. On the other hand,  more powerful state estimation techniques FF and LSTM are able to reconstruct the initial state even from a shorter sequence.

\begin{figure}
    \centering
    \includegraphics[width=.6\columnwidth]{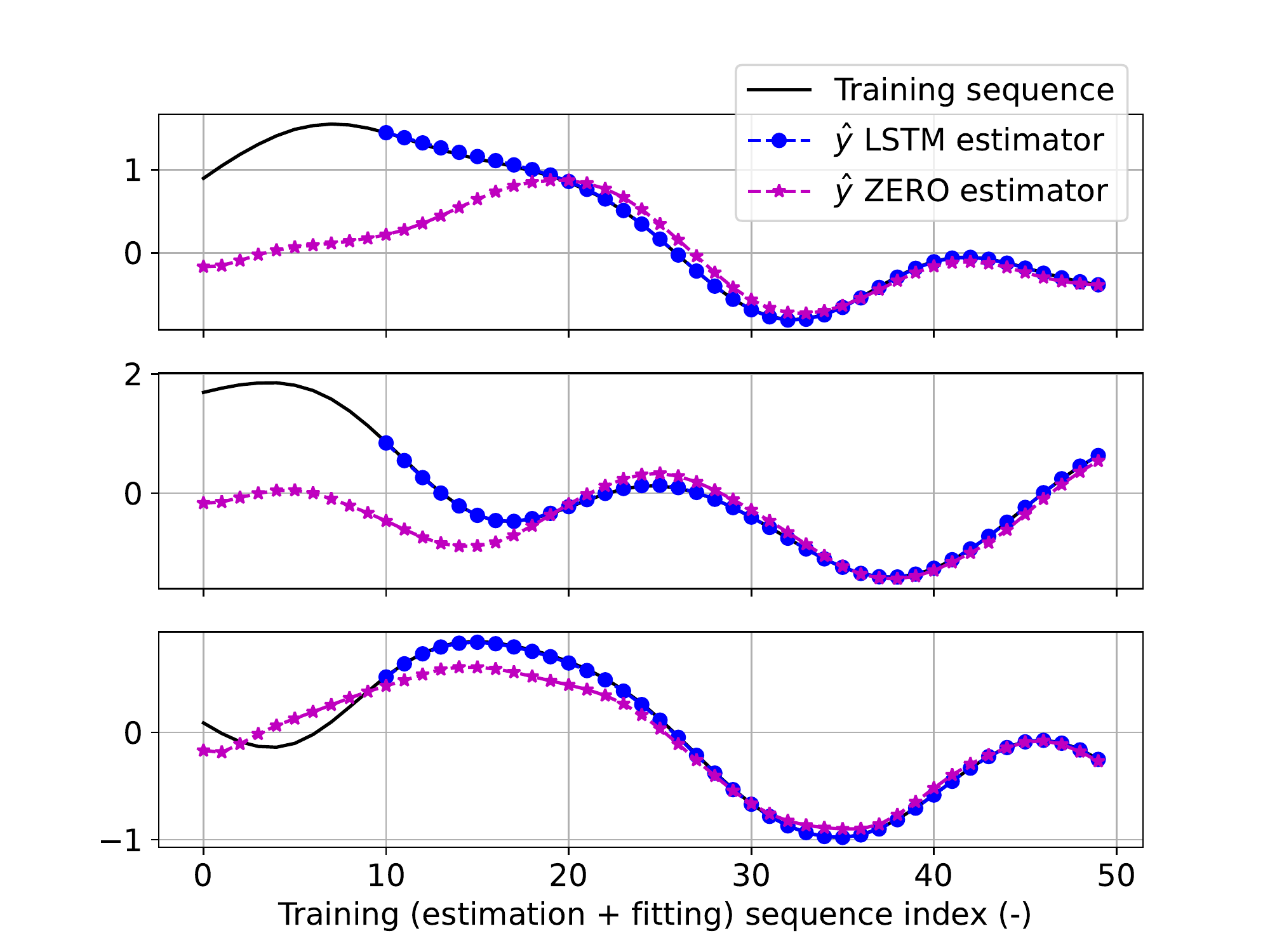}
    \caption{WH circuit: Training sequence (black) vs. model prediction obtained with est\_type=LSTM (blue), est\_type=ZERO (magenta). 
    The other factor are set to train\_time=3600 s, batch\_size=128, seq\_est\_len=10, seq\_len=40.}
    \label{fig:wh_subset_best_worst}
\end{figure}

The situation is highlighted in Fig. \ref{fig:wh_subset_best_worst}, where the behavior of the models obtained in two similar configurations 
(train\_time=3600 s, batch\_size=$\batchsize=128$, seq\_est\_len $\seqlen_e=10$, seq\_fit\_len $\seqlen_f=40$) and only differing in the estimator type (LSTM \emph{vs} ZERO) is reported for three random sequences extracted from the training dataset. In the LSTM case, the simulated output is already close to the target at time step 10, suggesting that this estimator is indeed effectively reconstructing the state at that time step. Conversely, the ZERO estimator (which is equivalent to a simulation initialized at time step 0 with state equal to 0) seems to require approximately 40 steps to converge to the true value, thus limiting the effectiveness of the overall approach. Incidentally, the plots also clarifies why, for configurations with seq\_est\_len $\seqlen_e \geq 40$, the fitting performance does not seem to
be affected by the choice of the state estimator in a significant way. Furthermore, for longer values of seq\_fit\_len $\seqlen_f$ (excluded in this specific analysis), the negative effect of bad state initialization is diluted over more fitting samples, and its impact on 
the model performance becomes negligible.

\begin{remark}
For smaller values of seq\_est\_len $\seqlen_e < 10$, the performance of the LSTM and FF state estimators may eventually decrease.
From a theoretical perspective, as already stated in Remark 
\ref{rem:min_seq_est_len_1}, even in a noise-free context a minimum number of samples is required for state estimation.
Specifically, a sufficient condition for exact state reconstructability is $\seqlen_e \geq \nx$ \cite{verdult2004least}.
The situation is more complex in a noisy scenario, and a detailed analysis is left as future work.
\end{remark}

\subsection{Pick-and-place machine}
In this case study, we consider data-driven modeling of an electronic component placement process in a pick-and-place machine. The experimental setup consists of a mounting head which holds an electronic component~\cite{juloski2004pickplace}. The component is pushed down to an impact surface simulating the printed circuit board and then it is released.   The system is characterized by two main operating modes: \emph{free} mode and \emph{impact} mode. In the free mode, the mounting head moves the component in an unconstrained environment without any contact with the impacting surface, while in the impact mode, the mounting head is in contact with the impacting surface. This setup is considered as a benchmark for assessing the effectiveness of \emph{hybrid} system identification approaches, see~\cite{bemporad2005bounded, piga2020bayes}.

An input-output dataset over a time interval of $15$~s with sampling frequency of $400$~Hz  is gathered. The training data consists of $5000$ samples taken during the first $12.5$ seconds of the experiment and the test dataset consists of $1000$ samples gathered in the last $2.5$ seconds.

We carry out an extensive experimental campaign in order to study the effects of the estimators  as well as different settings of the training algorithm on the identified model quality, quantified in terms of the {FIT} index \eqref{eq:fit_index} on the test dataset.

We consider the state-space neural model structure ~\eqref{eq:ss_model}  with order set to $\nx = 2$.
The neural networks $\NN_f$ and $\NN_g$ defining the state update and output are characterized by a single hidden layer with 15 nodes and $\mathrm{tanh}$ activation function. The networks are trained by employing \emph{Adam} algorithm with a learning rate set to $1\cdot 10^{-3}$.

We run the training algorithm for all the possible combinations of factors reported in  Table~\ref{tab:parameters_p_and_p} (full factorial experimentation) and thus we obtain $432$ models, each one corresponding to a different unique configuration. Execution of all the training runs required approximately 5 days.

\begin{table}[t!]
    \centering
    \begin{tabular}{|l||l|}
    \hline
   Factor &  Values \\
    \hline 
Estimator type (\texttt{est\_type}) & FF, LSTM, \\
& ZERO, RAND \\
\hline 
Training time (sec) (\texttt{max\_time}) & $300$, $1800$ \\
\hline 
Length $b$ of (mini)batch in training   (\texttt{batch\_size}) & $32$, $128$, $1032$ \\ 
\hline
Length $m_f$ of fitting sequence (\texttt{seq\_fit\_len})&  $64$, $256$, $512$ \\
\hline
Length $m_e$ of past I/O window (\texttt{seq\_est\_len}) &  $10$, $40$, $100$ \\
\hline
\# of hidden nodes of $\mathcal{N}_e$ (\texttt{est\_hidden\_size})&  $10$, $30$\\
\hline 
    \end{tabular}
    \vspace{0.2cm}
    \caption{Pick-and-place machine: Parameter configuration of the experiments.}
    \label{tab:parameters_p_and_p}
\end{table}

\begin{figure}
    \centering
    \includegraphics[width=.7\columnwidth]{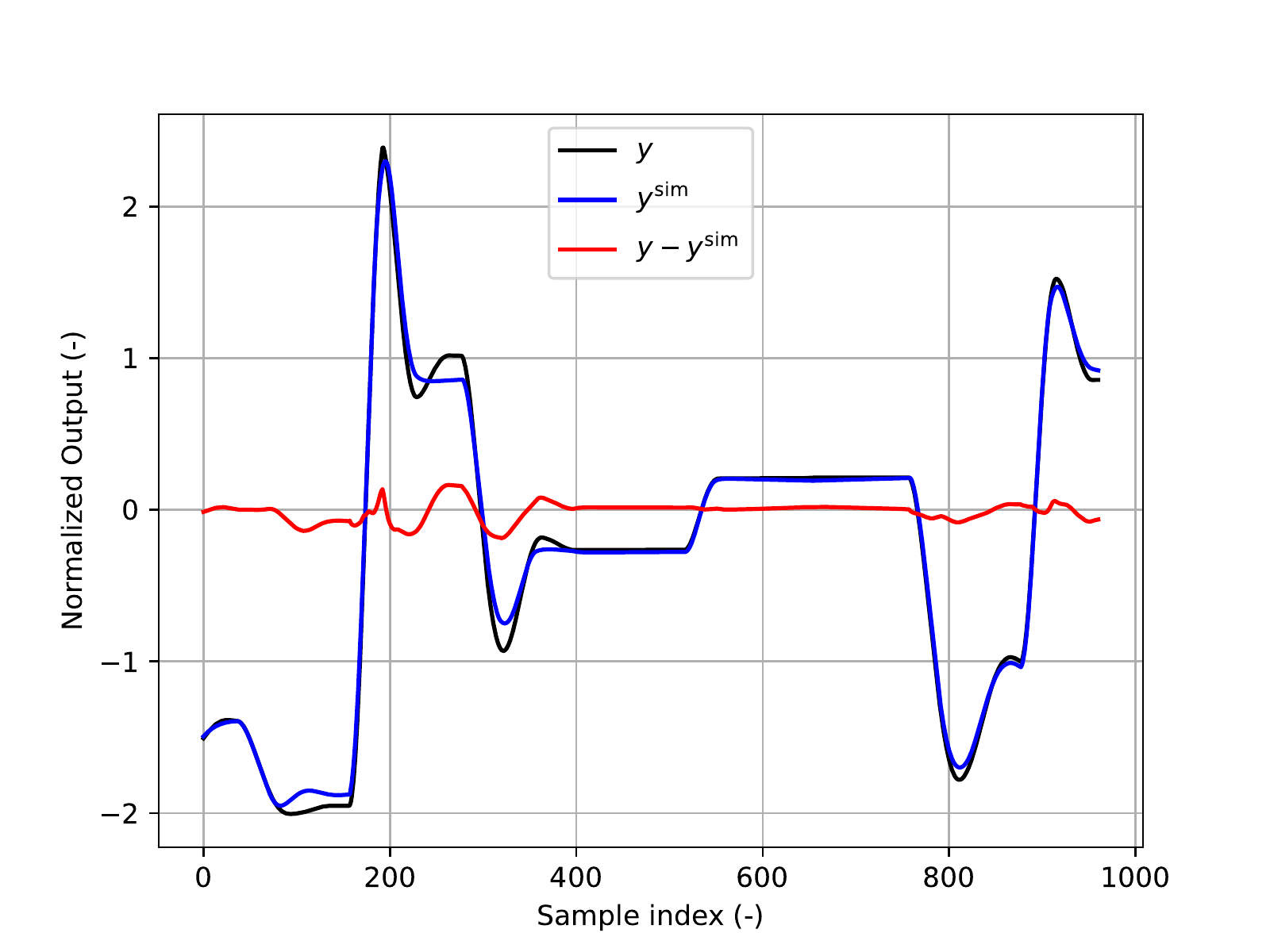}
    \caption{Pick-and-place machine: Measured output $y$, simulated output $y^{\rm sim}$, and error signal $y - y^{\rm sim}$ on the test dataset.}
    \label{fig:pp_best_timetrace.pdf}
\end{figure}

The top-$5$ configurations leading to the models with highest FIT index are reported in Table~\ref{tab:top_config_p_and_p}. It is interesting to note that, in contrast with the WH case study, all the top-5 configurations have non-dummy state estimators FF, LSTM. 
{To the best of our knowledge, the FIT indices  reported in Table~\ref{tab:top_config_p_and_p} are the highest achieved among the ones reported in the literature, see~\cite{bemporad2005bounded, piga2020bayes, mattson18, mejari2020mixed, mazzoleni22}  for this benchmark.}    
The measured output $y$, the output $y^{\rm sim}$ simulated from the best obtained model, and the error signal $y - y^{\rm sim}$ are shown in Fig.~\ref{fig:pp_best_timetrace.pdf}. 

\begin{figure}
    \centering
    \includegraphics[width=.7\columnwidth]{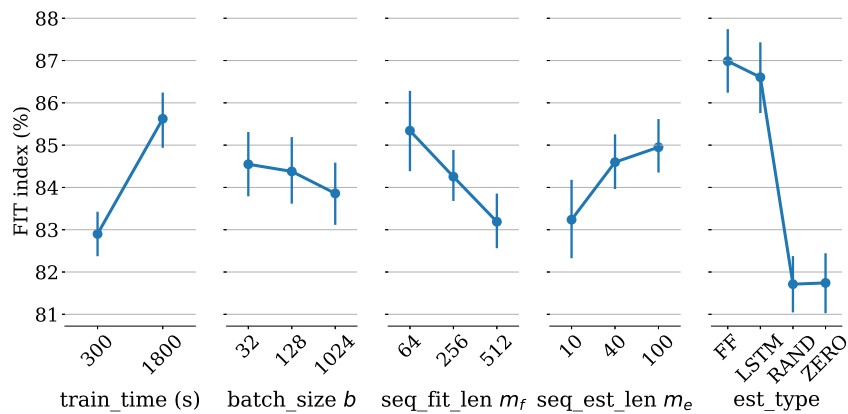}
    \caption{Pick-and-place machine: Effect of different factors on FIT index,  Mean (round dot) and $95\%$ confidence intervals (vertical bars). }
    \label{fig:fit_vs_fact_pnp}
\end{figure}

In Fig.~\ref{fig:fit_vs_fact_pnp}, we analyse the effect of different factors on the achieved model performance.\footnote{We omit to report results for the factor est\_hidden\_size, as its effect was found to be negligible.
} 

The round dot represents the mean value of the FIT index\footnote{For each value of a factor, the mean FIT is computed by taking the average over all the other factor values.} and the vertical bars represent 95\% confidence intervals.  In the leftmost subplot, we see that a larger training time ($1800$~s) gives better FIT on average w.r.t. a shorter training of $300$~s, as a consequence of the higher number of iterations allowed for convergence. The (mini)batch size $b$ has a slight negative effect on the FIT index, as seen in the second subplot. A longer fitting sequence length $\seqlen_f=512$ leads to worse performance, while a longer estimation sequence length $\seqlen_e$ improves the FIT on average, as seen from subplots $3$ and $4$ respectively. We remark that this simple evaluation of the effects of $\seqlen_f$ and $\seqlen_e$ on the FIT is not complete at this stage, as it is influenced by factors such as short training time, estimator type, etc. The effects of $\seqlen_f$ and $\seqlen_e$ will be analyzed in more details w.r.t. various factors in the subsequent paragraphs.

Finally, the effect of the state estimators on model quality can be seen in the rightmost subplot. This plot clearly shows that models having an estimator for the initial state (either FF or LSTM) significantly outperforms the models with ZERO or RAND initializations.  

\begin{remark}
	The significant improvement in the FIT index with FF and LSTM estimators is compatible with the following rationale: 
	We conjecture that for the pick-and-place machine dataset, the state estimators (FF, LSTM)  are required due to the presence of an integrator in the dynamics of underlying mechanical system. The  estimation error in the initial state may get integrated over time, resulting in a worse performance for models with RAND or ZERO initialization. On the other hand, FF and LSTM are able to provide an accurate estimate of the initial state. 
\end{remark}

\begin{table}[t!]
    \centering
    \begin{tabular}{|c|c|c|c|c|}
    \hline
    \texttt{est\_type}  & \texttt{batch\_size} & \texttt{seq\_fit\_len} & \texttt{seq\_est\_len} & \texttt{FIT} \\
    \hline
FF  & 128 & 64 & 40 & 93.61 \% \\
LSTM  & 32 & 64 & 100 & 93.40 \% \\
FF & 32 &  64 &40 & 92.94 \% \\
LSTM  & 128 & 64& 40 & 92.79 \% \\
LSTM & 128 & 64& 10 &  92.51 \% \\
\hline
\end{tabular}
\vspace{0.2cm}
    \caption{Pick-and-place machine: Top-$5$ configurations. For all configurations, train\_time = $1800$ s.}
    \label{tab:top_config_p_and_p}
\end{table}

\begin{figure}
	\centering
	\includegraphics[width=.7\columnwidth]{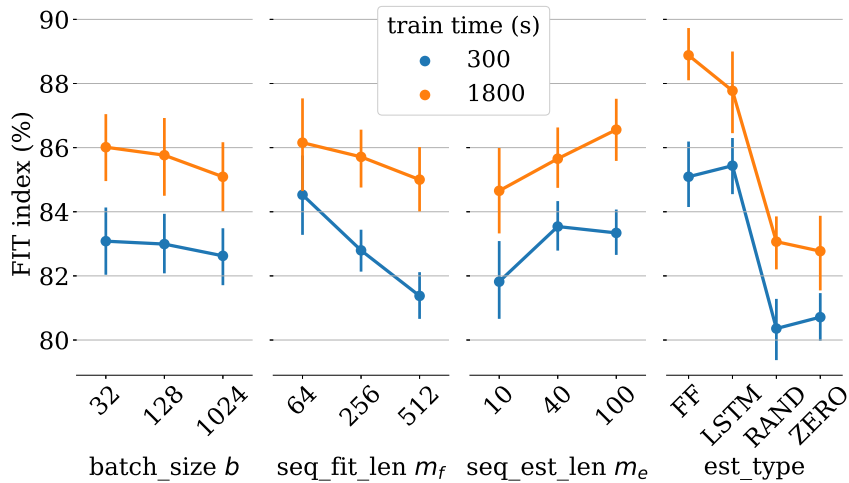}
	\caption{Pick-and-place machine: Effect of training time on FIT index.}
	\label{fig:PP_train_time}
\end{figure}

In Fig.~\ref{fig:PP_train_time}, the effect of the training time on FIT index is depicted. Naturally, a longer training time of $1800$~s implies a larger number of iterations allowing the algorithm to converge towards a better solution, which results in an improved model quality \emph{vs} models trained for a shorter interval of $300$~s. As discussed in the WH case study, for the short training time of $300$~s, it is interesting to note the trend of the FIT w.r.t. the fitting sequence length $\seqlen_f$  (blue curve shown in subplot 2). It can be seen that the model accuracy decreases fairly uniformly as $\seqlen_f$ is increased when models are trained for a short time interval. This phenomenon can be explained as follows: for a long fitting sequence, the training algorithm requires more time/iterations for convergence towards an optimum, thus pre-termination with a short training time results in models having generally lower performance. For a longer training time of $1800$~s, this effect is not prominent. From Fig.~\ref{fig:PP_train_time}, it is also interesting to note that the average performance is improved with increase in the length of the estimation sequence $\seqlen_e$ as seen in the third subplot from the left. 

\begin{figure}
	\centering
\begin{subfigure}[b]{.6\columnwidth}
	\includegraphics[width=\columnwidth]{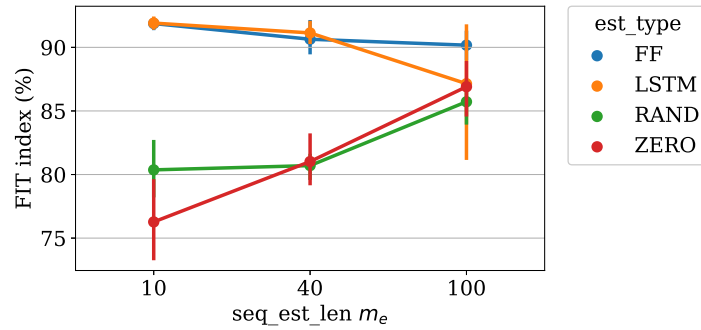}
	\caption{Effect of $m_e$ for fixed fitting sequence length $m_f =64$.}
	\label{fig:fit_seq}
\end{subfigure}
\begin{subfigure}[b]{.6\columnwidth}
	\includegraphics[width=\columnwidth]{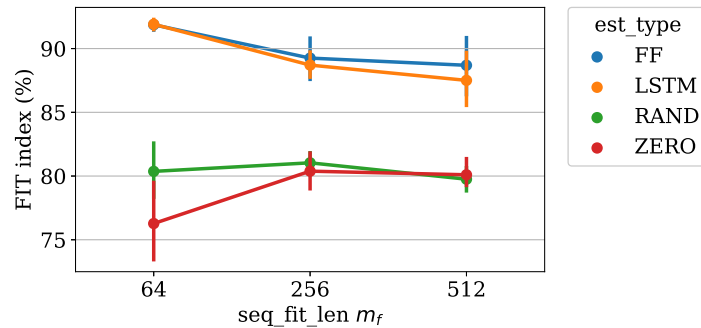}
	\caption{Effect of $m_f$ for fixed fitting sequence length $m_e =10$.}
	\label{fig:est_seq}
\end{subfigure}
\caption{Pick-and-place machine: Effect of $m_e$ and $m_f$ on FIT index with different estimator types.}
\label{fig:pp_fit_est_seq}
\end{figure}

Next, we focus on the effect of the factors seq\_fit\_len $\seqlen_f$ and seq\_est\_len $\seqlen_e$. In this analysis, we consider the larger training time of $1800$~s, omitting the results obtained with a short training interval of $300$~s. In Fig.~\ref{fig:pp_fit_est_seq}, we analyse the effect of {short} sequence lengths $\seqlen_e, \seqlen_f$ on the FIT index  for different estimators types. In particular, in Fig.~\ref{fig:fit_seq}, we plot the FIT index \emph{vs} $\seqlen_e$ with the seq\_fit\_len  $\seqlen_f= 64$, while in Fig.~\ref{fig:est_seq}, the FIT index is plotted against $\seqlen_f$ for a short estimation length of $\seqlen_e=10$. 
We observe that even for short estimation and fitting sequence lengths $\seqlen_e, \seqlen_f$, the estimators FF, LSTM result in a high FIT index. On the other hand, the FIT obtained with the dummy estimators ZERO, RAND is worse for short sequence lengths, with a high variability. Nonetheless,  it is interesting to note that for the dummy estimators ZERO, RAND, the FIT is improved for increasing values of $\seqlen_e$. 

\begin{figure}
	\centering
	\includegraphics[width=.5\columnwidth]{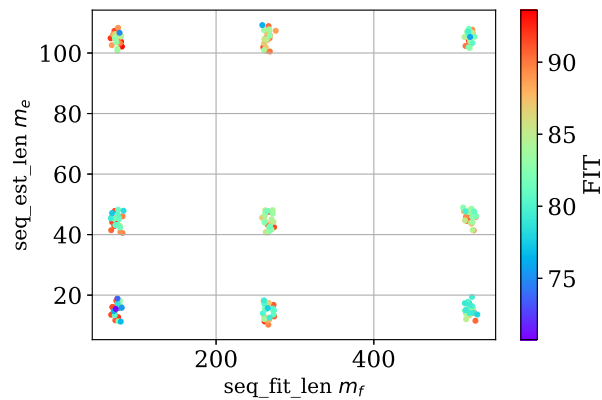}
	\caption{Pick-and-place machine: Effect of seq\_est\_len and seq\_fit\_len on FIT index for training time of $1800$~s.}
	\label{fig:pp_grid}
\end{figure}

In Fig.~\ref{fig:pp_grid}, we show the overall effect of different configurations of $\seqlen_e, \seqlen_f$ on the FIT index with different estimators. For the sake of visualization, the scatter plot is obtained by slightly perturbing $\seqlen_e, \seqlen_f$ (i.e., by adding some jitter to their numerical values on the grid, even though each cluster actually represents the same  configuration of $\seqlen_e, \seqlen_f$ having integer values). Similar to the previous analysis,  we observe that the lowest values of $\seqlen_e, \seqlen_f$ (i.e., lower left cluster with $\seqlen_e = 10, \seqlen_f = 64$) have the lowest average FIT index.  

\begin{figure}
	\centering
	\includegraphics[width=.7\columnwidth]{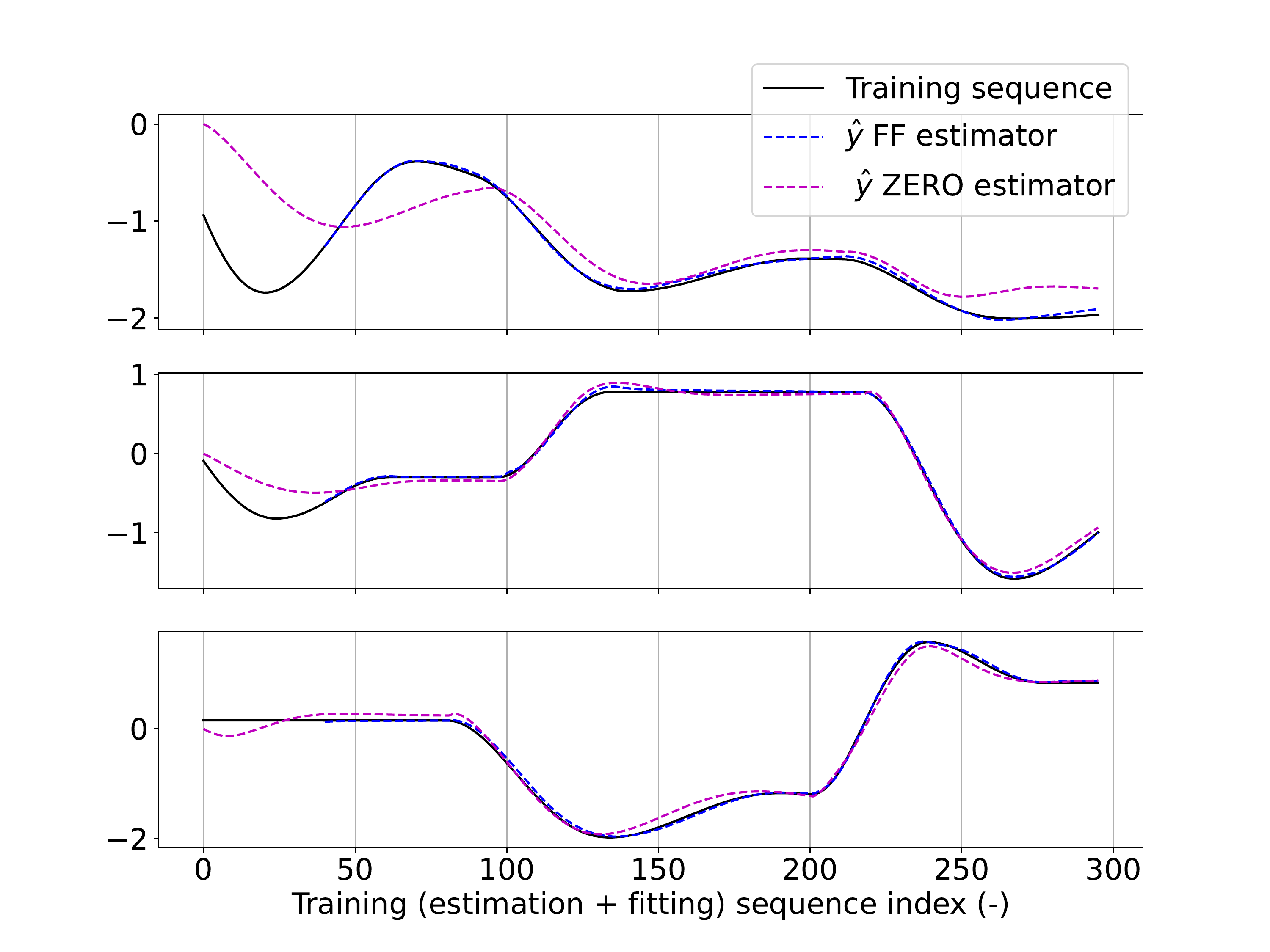}
	\caption{Pick-and-place machine: Training sequence (black) \emph{vs} model prediction obtained with est\_type=FF (blue), est\_type=ZERO (magenta). 
		The other factor are set to train\_time=1800 s, batch\_size=128, seq\_est\_len=40, seq\_len=256.}
	\label{fig:pp_subset_best_worst}
\end{figure}

Finally, we highlight the effect of ZERO \emph{vs} FF estimator on the model quality by visualizing the simulated outputs obtained for a specific training configuration.  In Fig.~\ref{fig:pp_subset_best_worst}, we plot $3$ different training sequences, along with the simulated output trajectories obtained with ZERO and FF estimator models, where other factors are fixed to $\batchsize = 128, \seqlen_e = 40, \seqlen_f = 256$. It is evident that the simulated trajectories obtained with the FF estimator follow the true sequence closely from the time step $40$, viz. the sequence length used to estimate the initial state. This suggests that the initial condition has been correctly reconstructed by the {FF} estimator. On the other hand, we observe that the simulated outputs with ZERO state initialization at time step $0$ requires about $100$ time steps to match closely with the true sequence, while still giving an error in the estimation at subsequent time samples. 
In particular, for the sequence in the first subplot which is further away from the zero initial condition, the mismatch between the trajectory simulated with ZERO initialization and the true sequence is large. This further highlights the need for an estimator for the pick-and-place benchmark example.  

Overall, the results obtained on the pick-and-place example demonstrate the need of a state estimator in order to significantly improve the model quality. In our assessment, this need is linked to the presence of an integrator in the underlying dynamics of the considered mechanical system. We conjecture that the overall lower FIT index obtained with ZERO and RAND initialization is a consequence of the integration of the initial state estimation error over time.

\section{Conclusion}
We have presented algorithms for neural state-space model learning and provided a detailed descriptions of their salient hyper-parameters and settings. Through extensive experimentation and analyses of the obtained results, we have extracted insight about the effects and interactions of several choices available to the system identification practitioner, focusing in particular on the state initialization procedure. 

The developed software has been made publicly available not only to allow reproducibility of the outcomes, but also to encourage other researchers to extend the analyses to additional factors, levels, and benchmarks. Possible investigation include: use of non-causal 
state reconstruction approaches; estimator design with formal strategies from control theory (e.g., Kalman, Luenberger, moving horizon); experimentation with unstable and chaotic dynamics.

We expect that similar studies could shed light on yet unknown properties of the existing algorithms, 
and possibly lead to improved and more efficient variations thereof.

\bibliography{references}
\bibliographystyle{plain}

\end{document}